\documentclass{article}
\usepackage{ijcai18}

\usepackage{times}
\usepackage{xcolor}
\usepackage{soul}
\usepackage[utf8]{inputenc}
\usepackage[small]{caption}
\usepackage{multirow}
\usepackage{amsmath}
\usepackage{graphicx}
\usepackage{epstopdf}
\usepackage{subfigure}

\title{Improving Deep Binary Embedding Networks by Order-aware Reweighting of Triplets}%\thanks{These match the formatting instructions of IJCAI-07. The support of IJCAI, Inc. is acknowledged.}}

\author{
Jikai Chen$^1$, 
Hanjiang Lai$^1$, 
Libing Geng$^1$, 
Yan Pan$^1$, 
\\ 
$^1$ Schoolf of Data and Computer Science, Sun Yat-sen University \\
}
\begin{document}

\maketitle

\begin{abstract} 
In this paper, we focus on triplet-based deep binary embedding networks for image retrieval task. The triplet loss has been shown to be most effective for the ranking problem. However, most of the previous works treat the triplets equally or select the hard triplets based on the loss. Such strategies do not consider the order relations, which is important for retrieval task. To this end, we propose an order-aware reweighting method to effectively train the triplet-based deep networks, which up-weights the important triplets and down-weights the uninformative triplets. First, we present the order-aware weighting factors to indicate the importance of the triplets, which depend on the rank order of binary codes. Then, we reshape the triplet loss to the squared triplet loss such that the loss function will put more weights on the important triplets. Extensive evaluations on four benchmark datasets show that the proposed method achieves significant performance compared with the state-of-the-art baselines.
%Our motivation is that since hashing is ranking problem, the order relations should also be considered to select triplets. 
%We adaptively weight t we reshape the triplet loss to the squared triplet loss such that the loss function will put more focus on training the hard triplets. he triplet via 1) the squared triplet loss to focus on the local order relation of the three binary codes and 2) the order-aware weighting factor by considering the order relations of the whole rank list. 
%We adaptively reweight the triplets based on how misranked via the current model both the order relation and the loss: 1) the order-aware weighting factors to indicate the importances of the triplets, which only depend on the rank order of binary codes. And 2) the squared triplet ranking loss to focus on hard triplets which have large losses. 

\end{abstract}

\section{Introduction}
With the rapid development of the Internet, the amount of images grows rapidly. The large-scale image retrieval has attracted increasing interest.  Hashing methods that encode images into binary codes have been widely studied since the compact binary codes are suitable for fast search and efficient storage. There are a multitude of hashing methods in the literature~\cite{wang2017survey,wang2016learning}. 
%which can be mainly divided into three categories: unsupervised methods~\cite{ITQ}, semi-supervised methods~\cite{SSH} and supervised methods~\cite{wang2016supervised,lin2017discriminative}. %Unsupervised methods learn the binary representation from the unlabeled training data. Semi-supervised methods take advantage of the information from both the labeled and unlabeled data. Supervised methods always have the best performances since they use the supervised information. 

Among these methods, the supervised information is given with triplet labels, which have been shown to be most effective since hashing is actually a ranking problem~\cite{zhuang2016fast,onestep}. In these works, the triplet ranking loss is defined to learn binary codes that preserve relative similarity relations.
%in the form of ``image $I$ is more similar to image $I^+$ than to image $I^-$". 
In \cite{onestep}, an architecture based on deep convolutional neural networks (CNNs) with triplet ranking loss is proposed for image retrieval. In \cite{zhao2015deep}, it presents a deep semantic ranking based method to learn hash functions that preserve multi-level semantic similarity between multi-label images. The FaceNet~\cite{schroff2015facenet} also uses the triplet ranking loss for face recognition and clustering. Due to the huge number of triplets, a collaborative two-stage approach~\cite{zhuang2016fast} is employed to reduce the training complexity of the triplet-based deep binary embedding networks. 
%A triplet ranking loss for face video retrieval is also proposed in \cite{dong2016face}.

%A major caveat of triplet-based deep network is that 
%The triplet loss relatively quickly learns to correctly preserve the similarities of the most trivial triplets, which makes a large fraction of all triplets uninformative~\cite{hermans2017defense}. FaceNet~\cite{schroff2015facenet} finds that the triplet loss converge quickly at the beginning, while after some point, the decrease of loss slows down drastically.

Not all triplets are of equal importance. In \cite{hermans2017defense}, it finds that the triplet loss relatively quickly learns to correctly map most trivial triplets, which makes a large fraction of all triplets uninformative after some point. Thus, the loss decreases quickly at the beginning and slows down drastically after some point~\cite{schroff2015facenet}.  For instance, the triplet $(bird_1, bird_2, dog_3)$ is easier than the triplet $(bird_1, bird_2, bird_3)$ in which three images are from the fine-grained bird database. Intuitively, the hash model which was told over and over again that bird and dog are dissimilar cannot further improve the performance. Hence, up-weighting the informative triplets and down-weighting the uninformative triplets become a crucial problem. 

However, most of the existing hashing methods treat the triplets equally~\cite{onestep}. Few works select the hard triplets based on the loss~\cite{wu2017sampling}. For instance, semi-hard negative mining~\cite{schroff2015facenet} is proposed in FaceNet. It uses all anchor-positive pairs in a mini-batch and selects the negative examplars that are further away from the anchor than the positive examplar. ~\cite{wang2015unsupervised} investigated to select top $K$ hard negative triplets with highest losses and the other triplets are ignored. \textit{In summary, all existing methods use the loss to select the hard triplets or treat them equally, and totally ignore the order relations in the rank list, which is important in retrieval task}. Since hashing problem is a ranking problem, the losses in the rank lists might not be sufficiently accurate than the order relations.

Inspired by that, we propose an order-aware reweighting method for triplet-based deep binary embedding networks, which up-weights the informative triplets and down-weights the uninformative triplets. We firstly introduce a weighting factor for each triplet. In practice, the weighting factor can be set to the value that indicates how the triplet is misranked by the current hash model. Hence, we use the MAP (\textit{mean average precision)}, which is a widely used evaluation measure, to calculate the weights. Specifically, for each mini-batch in the training phase, we encode the images into binary codes via deep CNNs. For an arbitrary triplet with an anchor, a positive code and a negative code, we rank all binary codes, including the positive code and the negative code, in the mini-batch according to their Hamming distances to the anchor. The weight of this triplet is defined as the change of MAP by only swapping the rank positions of the positive code and the negative code. Besides this order-aware weighting factor, we further use the squared triplet loss instead of the linear form, which up-weights hard triplets and down-weights easy ones from the perspective of the order relation of binary codes in triplets themselves. We conduct extensive evaluations on four benchmark datasets for image retrieval. The empirical results show that the proposed method achieves significant performance over the baseline methods.

\section{Related Work}

Hashing methods~\cite{wang2017survey} that learn similarity-preserving hash functions to encode data into binary codes have become popular methods for nearest neighbor search. Many methods have been proposed, which mainly can be divided into three categories: 1) the unsupervised hashing methods~\cite{shen2018unsupervised,liu2017reversed}, 2) the semi-supervised hashing methods~\cite{zhang2017ssdh,SSH} and 3) the supervised hashing methods~\cite{gui2018fast}. 

%e.g., spectral hashing (SH)~\cite{sh}, iterative quantization (ITQ)~\cite{ITQ}, anchor graph hashing (AGH)~\cite{AGH}, kernerlized LSH (KLSH)~\cite{KLSH} and semantic hashing~\cite{semantic}. 2) The semi-supervised hasing methods, e.g, semi-supervised hashing (SSH)~\cite{SSH} AND sequential projection learning for hashing (SPLH)~\cite{SPLH}. And 3) the supervised hashing methods, e.g., minimal loss hashing (MLH)~\cite{mlh},  kernel-based supervised hashing (KSH)~\cite{KSH} and binary reconstruction embedding (BRE)~\cite{BRE}. 

Learning the hash codes with deep frameworks, e.g., CNN-based methods~\cite{yang2018supervised}, has been emerged as one of the leading approaches. According to the forms of the supervised information, previous works mainly fall into three categories: 1) the point-wise approaches, 2) the pair-wise approaches and 3) the triplet-based/ranking-wised approaches. The point-wise methods take a single image as input and the loss function is built on individual data~\cite{lin2015deep}. %learns the hash codes and image representations in a point-wised manner. 
The pair-wise hashing methods take the image pairs as input and the loss functions are used to characterize the relationship (i.e., similar or dissimilar) between a pair of two images. For example, 
%CNNH~\cite{CNNH} is a two-stage hashing method based on pairwise labels. 
DPSH~\cite{li_pairwise} and DSH~\cite{liu2016deep} learn the hash codes by preserving the similarities among the input pairs of images. The triplet-based methods cast learning-to-hash as a ranking problem. \cite{onestep} proposed a deep triplet-based supervised hashing method. The triplet methods suffer from huge training complexity, thus \cite{zhuang2016fast} further proposed a two-step approach to accelerate the training process of triplet-based hashing network.

%hard example mining~\cite{lin2017focal} has been shown a great success in various applications including object detection~\cite{shrivastava2016training}, classification~\cite{loshchilov2015online} and so on. Some works~\cite{loshchilov2015online,wang2015unsupervised} select the hard examples based on the loss.
Recently, some works~\cite{wu2017sampling} show that sample selection plays an important role in learning the triplet-based network. The hard or semi-hard triplets are selected to train the network~\cite{wang2015unsupervised,schroff2015facenet}.  The distance weighted sampling~\cite{wu2017sampling} is proposed to select the informative and stable examples, where the samples are drawn uniformly according to their relative distance from one another. And the focal loss~\cite{lin2017focal} reshapes the standard cross entropy loss which down-weights the loss assigned to well-classified examples. %The proposed focal loss can prevent the huge number of easy examples overwhelming the detector during training. 

%In \cite{wang2015unsupervised}, it finds hard examples that have highest losses. \cite{shrivastava2016training} proposed online hard example mining for image detection problem. There are overwhelming number of easy examples and a small number of hard examples in the dataset, and automatic select the hard examples can make training more effective and efficient. Further, \cite{lin2017focal} reshaped the standard cross entropy loss which down-weights the loss assigned to well-classified examples. The proposed focal loss can prevent the huge number of easy examples overwhelming the detector during training. 

Inspired by these methods, we propose an order-aware method to reweight the triplet loss. The existing methods use the loss to select the hard examples or treat the triplets equally. In contrast, our proposed method introduces the order information~\cite{donmez2009local} to weight the triplets, which is much more effective and accurate.

\section{Overview of Triplet-based Hashing Networks}

\begin{figure}
  \centering
    \includegraphics[width=1\hsize \hspace{0.01\hsize}]{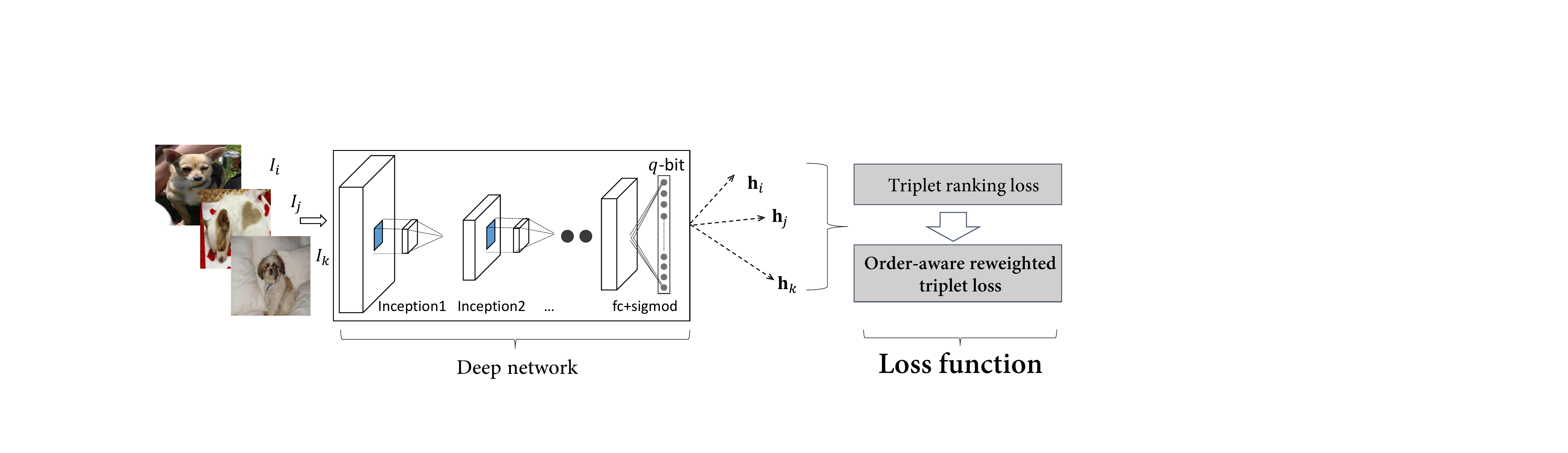}
  \caption{Overview of the triplet-based hashing network. The triplet-based network consists two sequential parts: a deep network and a triplet loss. In this paper, we only focus on the loss function. We reshape the triplet loss to our order-aware reweighted triplet loss.}
  \label{example_network}  % be used to cite the whole fig
\end{figure}

In this section, we briefly summarize the triplet-based hashing framework. It takes triplets of images as inputs, i.e., $(I_i, I_j, I_k)$, in which $I_i$ is semantically more similar to $I_j$ than to $I_k$. The triplet hashing network itself can be divided into two sequential parts: a deep network with a stack of convolution, max-pooling and fully-connected layers; and a triplet ranking loss layer as shown in Figure~\ref{example_network}. 

In deep network, the convolutional layers are applied to produce powerful feature maps, which encode the images into high-level representations. Then the following several fully-connected layers project the feature maps into the desired-length feature vectors, e.g., $q$-dimensional vectors, where $q$ is the length of binary codes. The feature vectors are fed into a sigmoid layer which is smooth and well approximated the threshold function. The outputs of the network are restricted in the range $[0,1]^q$. We denote the outputs of triplet network as $\mathbf{h}_i = \mathcal{F}(I_i)$, where $I_i$ is the input image and $\mathcal{F}$ is the deep network.

Through the deep network, triplet ranking loss~\cite{onestep} is used to preserve the relative similarities of images. Given the input images in the form of $(I_i, I_j, I_k)$, the goal of hash network is to preserve the similarities of the learned binary codes, i.e., the binary code $\mathbf{h}_i$ is closer to $\mathbf{h}_j$ than to $\mathbf{h}_k$. The triplet ranking loss is defined by
\begin{equation}
\label{equation_triplet}
\begin{split}
& \ell_{(i,j,k)} =  \ell (\mathbf{h}_i,\mathbf{h}_j,\mathbf{h}_k) \\
= & \text{max}(0, \epsilon - ||\mathbf{h}_i-\mathbf{h}_k||_2^2 + ||\mathbf{h}_i-\mathbf{h}_j||_2^2),  \\
\end{split}  
\end{equation}
where $\epsilon$ is a hyper-parameter to control the  margin between the two distances. %$\lVert\cdot \rVert _2$ denotes $l_2$ norm. 

%The number of triplets scales cubically with the number of training images. However, there are an overwhelming number of easy triplets and a small number of hard triplets. The following is an intuitive example. After 10 epochs of training using randomly selected triplets, we save the hash model. Note that this hash model is far from converging to the optimal solution. Then, this model is used to calculate the triplet loss in E.q.~\ref{equation_triplet}. As shown in Fig.~\ref{}, the dots are the losses of triplets, respectively. We can seen that the hash model can preserve the similarities for the easy samples while there are still some hard examples that have large errors. 

%In training, it is expected that the hard triplets can be selected more. Unfortunately, if we treat the triplets equally, the contributions of uninformative triplets may dominate the overall loss when there are overwhelming number of easy triplets, and the important triplets have little impact on the triplet ranking loss function. 

% Here, we randomly select 500 images on SUN397 dataset and construct the triplets according to their labels.

%The main problem of triplet loss: \textit{vast number of easy examples and sparse set of hard examples}. (draw a figure to show the large of easy examples and few hard examples). 

\section{Order-aware Reweighting of Triplets}
In this section, we only focus on the loss function and propose a simple yet effective order-aware reweighting algorithm for image retrieval. We first introduce the motivation of this work and then elaborate the proposed method. 
% We discover that there are extreme easy-hard triplets imbalance encountered in the triplet ranking loss, which will result in suboptimal solution. 

\subsection{Motivation}

\begin{figure}[ht!]
  \centering
    \includegraphics[width=0.9\hsize \hspace{0.01\hsize}]{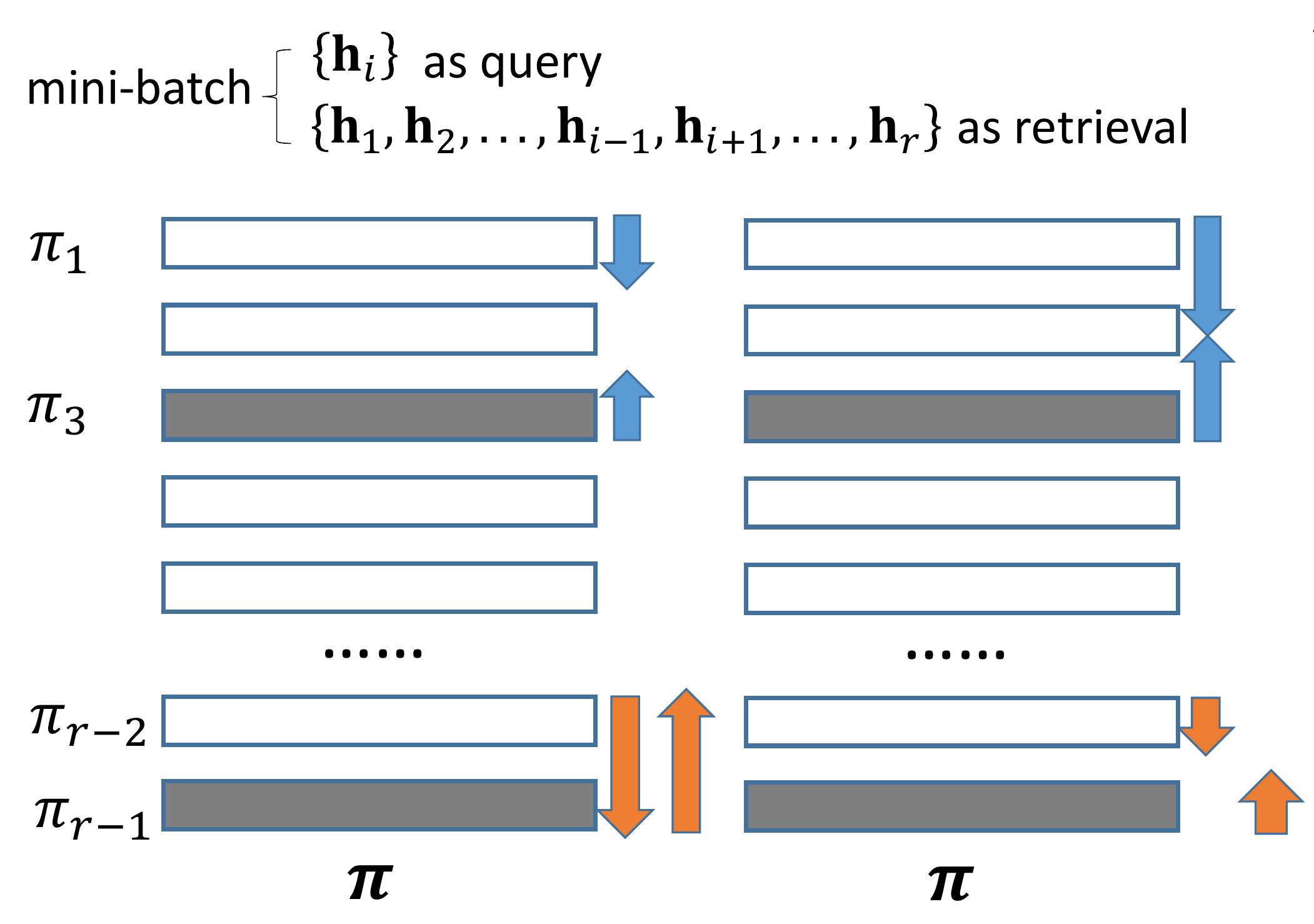}
  \caption{%Motivation of our method. Given a query code $q$, we return a rank list $\pi$. The white bars represent codes that are irrelevant to the query, and black bars represent the relevant codes. Left: the arrows denote the triplet ranking losses on two triplets $(\mathbf{q},\mathbf{h}_{\pi_3},\mathbf{h}_{\pi_1})$ and $(\mathbf{q},\mathbf{h}_{\pi_n},\mathbf{h}_{\pi_{n-1}})$. If we only consider the loss, the triplet $(\mathbf{q},\mathbf{h}_{\pi_n},\mathbf{h}_{\pi_{n-1}})$ will be assigned more weight. However, for the ranking problem, we would really like to move $\mathbf{h}_{\pi_3}$ to the top 1 position but not give heavy weight to $\pi_n$ (since the top positions are more important than the bottom positions). Right: the better choice of the proposed order-aware loss function. 
  Motivation of our method. Given a query code $\mathbf{h}_i$ in a mini-batch, we return a rank list $\pi$ for $\mathbf{h}_i$. The white bars represent codes that are irrelevant to the query, and black bars represent the relevant codes. The arrows denote the weights and the moving directions of the items. Two example triplets, $\ell(\mathbf{h}_i,\mathbf{h}_{\pi_3},\mathbf{h}_{\pi_1}) = 1$ and $\ell(\mathbf{h}_i,\mathbf{h}_{\pi_{r-1}},\mathbf{h}_{\pi_{r-2}}) = 4$, are misranked by the current hashing model. Left: the weights of the triplets (see the arrows on the left) based on the loss. However, the $\mathbf{h}_{\pi_3},\mathbf{h}_{\pi_1}$ are in the top positions and top results are more important in retrieval task. To better quantify the misranked triplet, we adopt the change of MAP by swapping the positions of the positive and negative codes to weight these misranked triplets. Right: the better choice to reweight the two triplets (see the arrows on the right). }
  \label{example_wapr}  % be used to cite the whole fig
\end{figure}

In retrieval task, the misranked triplets $(\mathbf{h}_i,\mathbf{h}_j,\mathbf{h}_k)$ in which $\mathbf{h}_j$ ranks behind  $\mathbf{h}_k$ when $\mathbf{h}_i$ is query, should be further emphasized to boost the model. A simple and intuitive method is to add more weights to these misranked triplets. However, existing methods treat the triplets equally or only use the loss to select the hard ones. Image hashing is a ranking problem. Thus, the order relations are desirable. Figure~\ref{example_wapr} is an example illustration. Given $\mathbf{h}_i$ as the query, we rank the other binary codes according to their Hamming distance to $\mathbf{h}_i$ and $\pi = \{\pi_1,\cdots,\pi_{r-1}\}$ is the returned rank list. Take two misranked triplets $(\mathbf{h}_i,\mathbf{h}_{\pi_3},\mathbf{h}_{\pi_1})$ and $(\mathbf{h}_i,\mathbf{h}_{\pi_{r-1}},\mathbf{h}_{\pi_{r-2}})$ as examples in which $\ell(\mathbf{h}_i,\mathbf{h}_{\pi_3},\mathbf{h}_{\pi_1})=1$ and $ \ell(\mathbf{h}_i,\mathbf{h}_{\pi_{r-1}},\mathbf{h}_{\pi_{r-2}})=4$, if we only
use the loss to weight the triplets, the triplet $(\mathbf{h}_i,\mathbf{h}_{\pi_{r-1}},\mathbf{h}_{\pi_{r-2}})$ will have larger weight. However, since $\mathbf{h}_{\pi_{1}}$, $\mathbf{h}_{\pi_{3}}$ are in the top positions and the top items are more important for the retrieval task, simply swapping the positions of $\mathbf{h}_{\pi_{1}}$ and $\mathbf{h}_{\pi_{3}}$ can achieve better performance than simply swapping those of $\mathbf{h}_{\pi_{r-2}}$ and $\mathbf{h}_{\pi_{r-1}}$. Hence, the triplet $(\mathbf{h}_i,\mathbf{h}_{\pi_3},\mathbf{h}_{\pi_1})$ should be assigned more weight than $(\mathbf{h}_i,\mathbf{h}_{\pi_{r-1}},\mathbf{h}_{\pi_{r-2}})$. The better choice is shown in the right side of Figure~\ref{example_wapr}. 

In this paper, we propose a simple algorithm that down-weights the uninformative triplets and up-weights the informative triplets via  1) a order-aware weighting factor and 2) a squared triplet ranking loss.

\subsection{Order-aware Weighting Factor}
 
Observed by that, we add an order-aware weighting factor to indicate the importance of the triplet. If the triplet is important, the more weight should be assigned to this triplet. We use the following steps to obtain the order-aware weights for triplets.

(1) Triplet generation with order information. 
Given a mini-batch of $r$ images, these images go through the deep network and are encoded as $\mathbf{h}_i, i=1,\cdots,r$. Then we construct $r$ rank lists $\pi^{(1)},\pi^{(2)},\dots,\pi^{(r)}$. The $i$-th rank list is constructed for the $i$-th binary code, in which given the $i$-th code as the query, we rank the other $r-1$ codes according to their Hamming distance to the $i$-th code, e.g., $\pi^{(i)} = \{\pi^{(i)}_1, \cdots, \pi^{(i)}_{r-1} | D_H(\mathbf{h}_i,\mathbf{h}_{\pi^{(i)}_1}) \leq \cdots \leq D_H(\mathbf{h}_i,\mathbf{h}_{\pi^{(i)}_{r-1}}) \}$, where $D_H(\cdot,\cdot)$ denotes Hamming distance function. With these rank lists, we generate the set of triplets for the $i$-th query code: $T^{(i)} = \{(\mathbf{h}_i, \mathbf{h}_{\pi^{(i)}_j}, \mathbf{h}_{\pi^{(i)}_k}) | sim(\mathbf{h}_i, \mathbf{h}_{\pi^{(i)}_j}) > sim(\mathbf{h}_i,\mathbf{h}_{\pi^{(i)}_k}) \}$ where $sim(\cdot,\cdot)$ is semantic similarity measure. Note that the first item is always $\mathbf{h}_i$ in the set $T^{(i)}$. Then, the union of all the $r$ sets, $T = \bigcup_{i=1}^{r} T^{(i)}$, is the total set of all triplets in the mini-batch. 

%For each binary code $\mathbf{h}_i$, the triplets of $h_i$ are generated according to the supervised information in mini-batch: $T = \{(h_i, h^+, h^-) | sim(h_i, h^+) > sim(h_i,h^-) \}$ where $sim(\cdot,\cdot)$ is semantic similarity measure. Now we show how to define the weighting factor for these triplets. 

(2) Order-aware reweighting of triplets.
Now the problem becomes how to define weighting factors for these triplets. Take the set $T^{(i)}$ as an example, given a triplet $(\mathbf{h}_i, \mathbf{h}_{\pi^{(i)}_j}, \mathbf{h}_{\pi^{(i)}_k})$, we denote $\lambda_{(i,j,k)}$ as the importance weight of this triplet. Since the MAP is a widely used evaluation measure for ranking, we adopt MAP to calculate the weights. More specifically, for the triplet $(\mathbf{h}_i, \mathbf{h}_{\pi^{(i)}_j}, \mathbf{h}_{\pi^{(i)}_k})$, we first calculate the MAP of the rank list $\pi^{(i)}$ for the query $\mathbf{h}_i$. Then we only swap the rank positions of $\pi^{(i)}_j$ and $\pi^{(i)}_k$, and the other rank positions are fixed in rank list $\pi^{(i)}$, through which we can obtain another MAP. The absolute value of the difference between the two MAPs is used as the weight of the triplet. More specifically, let $\pi^{(i)} = \{ \cdots, \pi^{(i)}_k, \cdots, \pi^{(i)}_j, \cdots  \}$ and $\hat{\pi}^{(i)} = \{ \cdots, \pi^{(i)}_j, \cdots, \pi^{(i)}_k, \cdots \}$. Note that other positions in the two rank lists $\pi^{(i)}$ and $\hat{\pi}^{(i)}$ are the same,  and only the rank positions of $\pi^{(i)}_j$ and $\pi^{(i)}_k$ are swapped. The order-aware weight for the triplet $(\mathbf{h}_i, \mathbf{h}_{\pi^{(i)}_j}, \mathbf{h}_{\pi^{(i)}_k})$ is defined as 

%we rank other $r-1$ codes according to the distances to the $h_i$. A rank list is obtained for $h_i$ : $\{h_{\pi_1},\cdots,h_{\pi_{r-1}}\}$. The order-aware weighting factor of $(h_i, h^+, h^-)$ is defined as "the change in evaluation measures by swapping the rank position of $h^+$ and $h^-$", (draw a figure ...) which is defined as
\begin{equation}
\lambda_{(i,j,k)} = | MAP (\pi^{(i)}) - MAP(\hat{\pi}^{(i)}) |.
\end{equation}
 
\subsection{Squared Triplet Ranking Loss}
 
% \begin{figure}[!t]
%\includegraphics[width=3.5in]{plt.jpg}
%\caption{When $\gamma$ larger than 1, the loss of hard triplets with larger loss up-weights and easy triplets with smaller loss down-weights after transformation.}
%\end{figure}

%As discussed in the previous section, a large number of easy triplets comprise the majority of the loss and dominate the gradient when back propagation is applied to update the weights of the deep networks. Take 10 triplets as an example, there are one hard triplet, e.g., $\ell_1 = 5$ and nine easy triplets, e.g., $\ell_i = 1, i=2,\cdots,10$. The overall loss is 14, in which the loss of the easy triplets overwhelm that of the hard triplet. A simple and intuitive method for increasing the weights of hard triplet is to reshape the loss from linear function to quadratic function. By using the quadratic function, we have $[\ell_1]^2 = 25$ and $[\ell_i]^2 = 1$ for other nine easy triplets. Thus, the loss of hard triplet can dominate the overall loss. 

Instead of the linear function, we propose a squared triplet ranking loss which aims to down-weight uninformative triplets and focuses on training hard distinguished triplets. The triplet loss function is changed as:
\begin{equation}
\ell_{(i,j,k)} \to [\ell_{(i,j,k)}]^2.
\label{squared_loss}
\end{equation}
%When the loss of a triplet is small, the quadratic error is also small. As the loss increases, the quadratic loss increases faster.  This in turn reduces the importances of the these easy triplets and increases the importances of the hard triplets. 

By using the squared triplet loss, the informative triplets will comprise the majority of the loss and dominate the gradient when back propagation is applied to update the weights of the deep networks. Even for the case that there are overwhelming number of uninformative triplets and a small number of informative triplets. Take 10 triplets as an example, there are one harder triplet, e.g., $\ell_1 = 5$ and nine easier triplets, e.g., $\ell_i = 1, i=2,\cdots,10$. The overall loss is 14, in which the loss of the easy triplets overwhelms that of the hard triplet. 
%A simple and intuitive method for increasing the weights of the hard triplet is to reshape the loss from linear function to quadratic function. 
By using the quadratic function, we have $[\ell_1]^2 = 25$ and $[\ell_i]^2 = 1$ for other nine easy triplets. Thus, the loss of hard triplets can dominate the overall loss~\footnote{Note that the conclusion also can be obtained when the losses are less than one, e.g., $\ell_1 = 0.5$ and $\ell_i = 0.1, i=2,\cdots,10$.}. 

\begin{figure*}[ht!]
%\tiny
%Requires \usepackage{graphicx}
  \begin{flushleft}
  \centering
  \subfigure[VOC2007]{\label{voc-pr}
   \raisebox{-0.01cm}{%
   \includegraphics[width=0.22\textwidth]{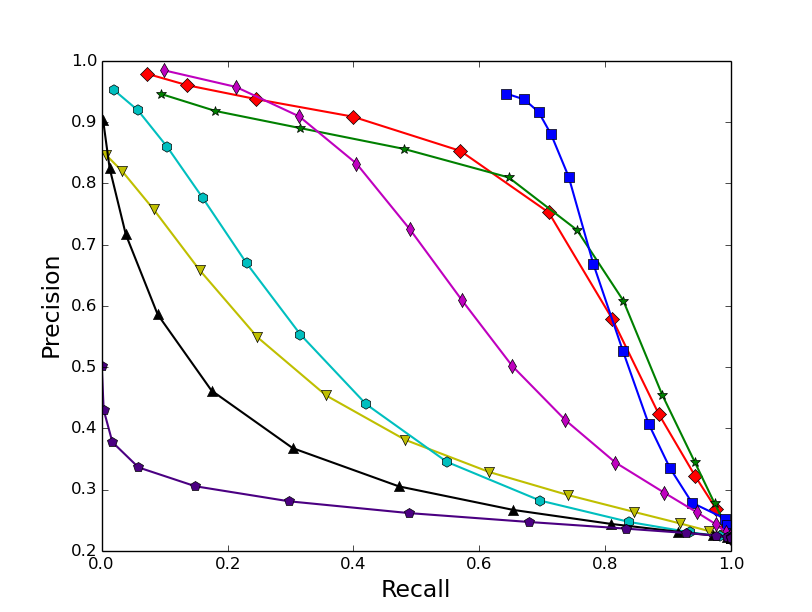}
  }}
   \subfigure[Stanford Dogs]{\label{dog-pr}
   \raisebox{-0.01cm}{
   \includegraphics[width=0.22\textwidth]{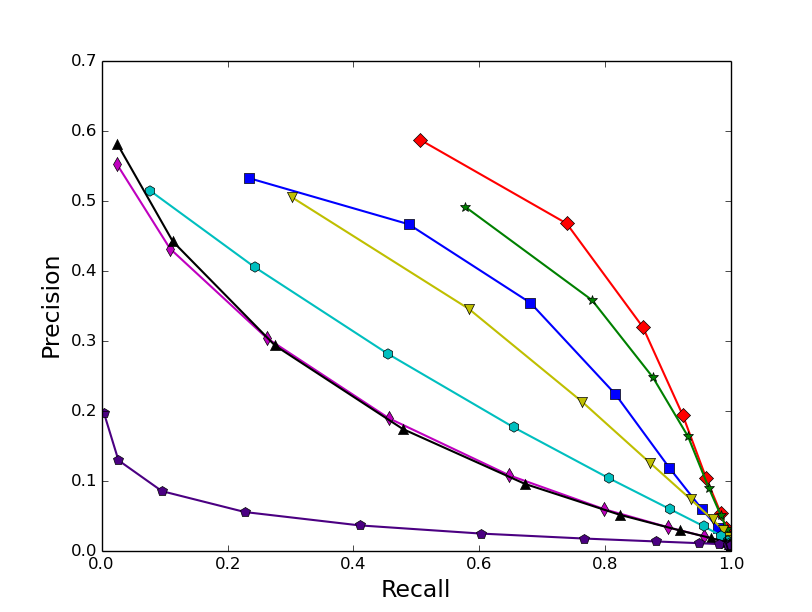}
  }}
   \subfigure[SUN397]{\label{sun-pr}
   \raisebox{-0.01cm}{
   \includegraphics[width=0.22\textwidth]{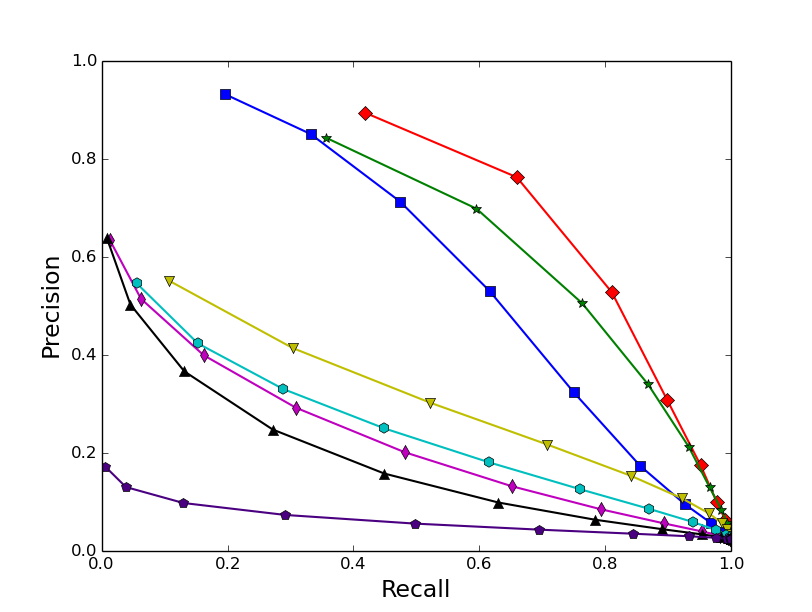}
  }}
  \subfigure[CUB-200-2011]{\label{cub-pr}
  \raisebox{-0.01cm}{\includegraphics[width=0.22\textwidth]{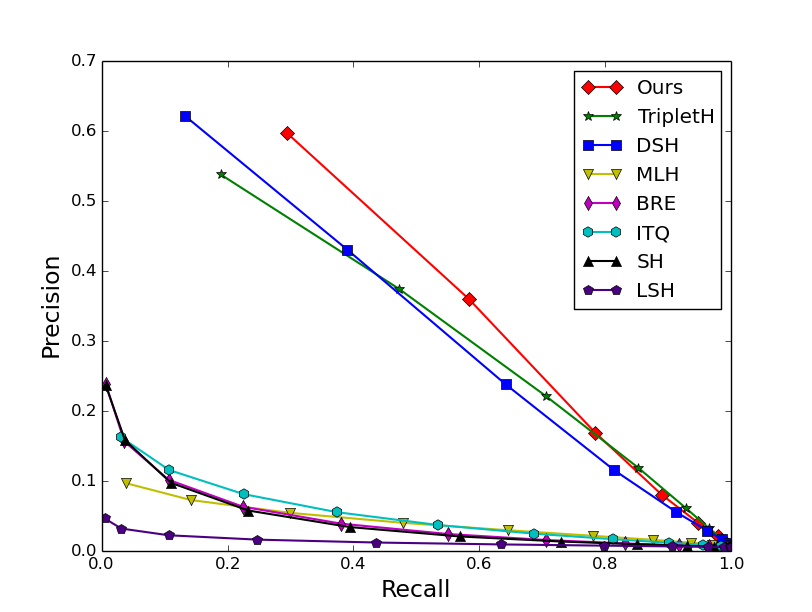}
  }}
  \caption{Precision-recall curves}
  \label{pr-result}
  \end{flushleft}
\end{figure*}

\begin{figure*}[ht!]
%\tiny
%Requires \usepackage{graphicx}
  \begin{flushleft}
  \centering
  \subfigure[VOC2007]{\label{voc-p}
   \raisebox{-0.01cm}{%
   \includegraphics[width=0.22\textwidth]{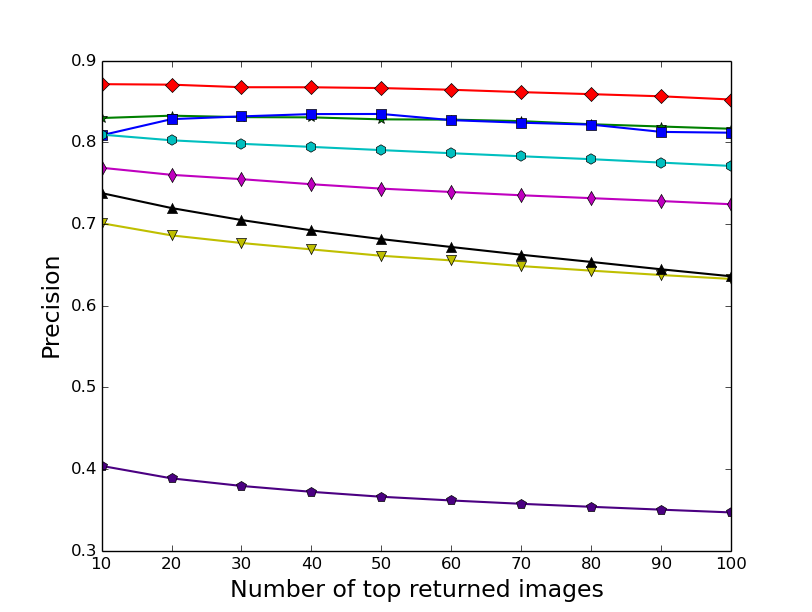}
  }}
   \subfigure[Stanford Dogs]{\label{dog-p}
   \raisebox{-0.01cm}{
   \includegraphics[width=0.22\textwidth]{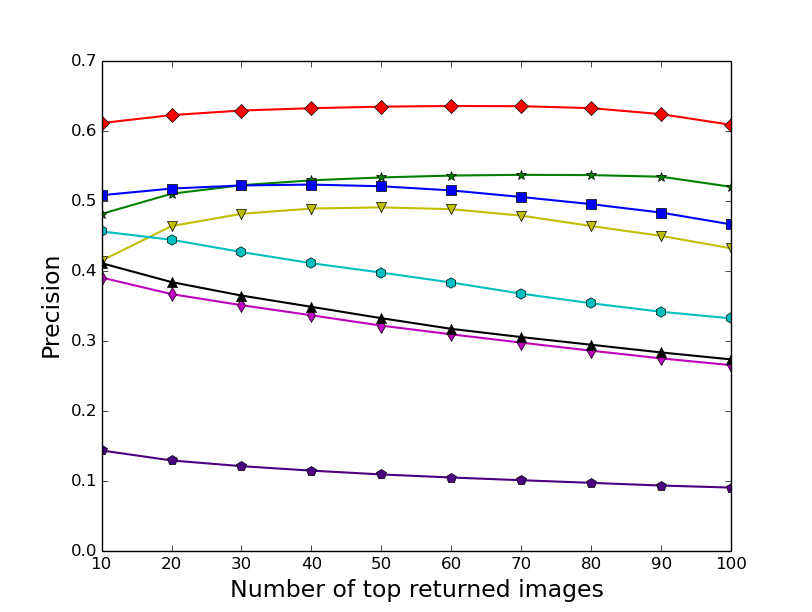}
  }}
   \subfigure[SUN397]{\label{sun-p}
   \raisebox{-0.01cm}{
   \includegraphics[width=0.22\textwidth]{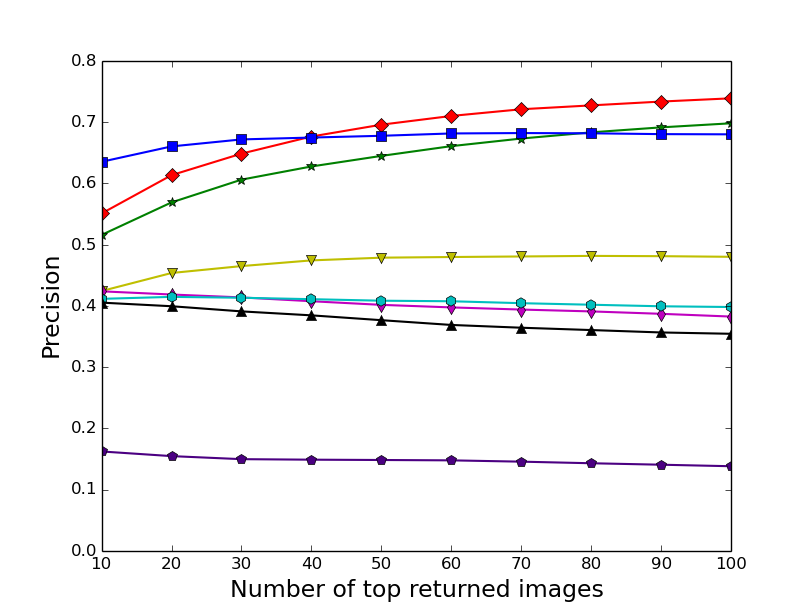}
  }}
  \subfigure[CUB-200-2011]{\label{cub-p}
  \raisebox{-0.01cm}{\includegraphics[width=0.22\textwidth]{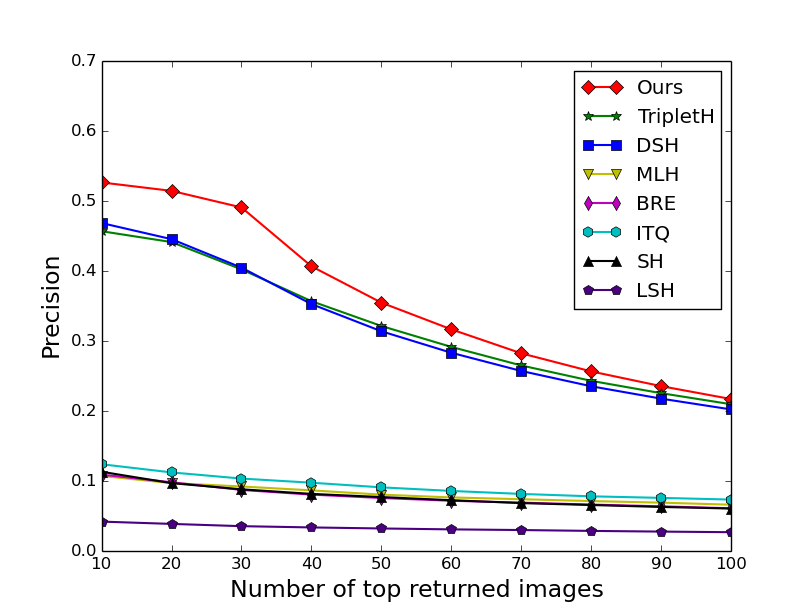}
  }}
  \caption{Precision curves}
  \label{p-result}
  \end{flushleft}
\end{figure*}

\subsection{Full Objective}
The overall objective for order-aware reweighting triplet loss is defined as
\begin{equation}
\min \sum \lambda_{(i,j,k)} \times [\ell_{(i,j,k)}]^2.
\end{equation} 

Our loss function contains two terms: 1) the order-aware weighting factors, which define the importance of the triplets by considering the whole rank list. It will let the loss function focus on the triplets that have the worst rank positions. And 2) the squared ranking objective is to up-weight the hard triplets by considering the order relation of binary codes in triplets.

\section{Experiments}
\subsection{Datasets and Evaluation Measures}
In this section, we conduct extensive evaluations of the proposed method on four benchmark datasets:

\begin{itemize}

\item{\textbf{VOC2007}}~\cite{voc2017}: It consists of 9,963 annotated consumer photographs collected from the Flickr~\footnote{http://www.flickr.com} photo-sharing website. There are 20 object classes in this dataset, and each image is annotated with 1.5 labels on average.
 
\item{\textbf{Stanford Dogs}}~\cite{KhoslaYaoJayadevaprakashFeiFei_FGVC2011}: It contains 20,580 images of 120 breeds of dogs from around the world, which has been built using images and annotation from ImageNet for the task of fine-grained image categorization. 

\item{\textbf{SUN397}}~\cite{xiao2010sun}:  It contains 397 scene categories. The number of images varies across categories, but there are at least 100 images per category and 108,754 images in total.

\item{\textbf{CUB-200-2011}}~\cite{wah2011caltech}: It is a challenging dataset of 200 bird species. All 11,788 images and annotations were filtered by multiple users of Mechanical Turk.

\end{itemize}

In VOC2007, Stanford Dogs and CUB-200-2011 three datasets, we utilize the official train/test partitions to construct the query sets and the retrieval databases. The testing samples are used as the query set, and the rest images, i.e., the training samples, are used as the retrieval database. The training samples are also used to train the hash models. Note that the validation set of VOC2007 is included in the retrieval database but not used in training.
                                                                                                   
In SUN397 dataset, we follow the setting of~\cite{do2016binary} and use the subset of images associated with the 42 categories with each containing more than 500 images~\footnote{In the official train/test partition, they use a subset of dataset that has 50 training images and 50 testing images per class, averaging over the 10 partitions. It is not suitable for hashing problem.}. The randomly sampled 4,200 images (100 images per class) are constructed as the query set. The rest images are used as the database for retrieval. We randomly select 400 samples per class from the retrieval database to form the training set.

To evaluate the quality of hashing, we use the following evaluation metrics: mean average precision (MAP), precision-recall curves, and precision curves w.r.t. different numbers of top returned samples. 

%\textbf{Evaluation Measures}. MAP is a widely used evaluation measure for ranking.  The average precision of image $x_i$ is defined as the average of all
%$P_i@k (k=1,\cdots,c)$ and can be calculate by
%\begin{equation}
%	 AP_i = \sum_{k=1}^c P_i@k \times pos(k)/ |c_{+}^i|,
%\end{equation}
%where $k$ is the position, $c$ is the number of labels, and $pos(k)$ is an indicator function, in which if the label at position $k$ is positive, then $pos(k) = 1$, or else $pos(k) = 0$.
%
%For all images, the MAP is defined as
%\begin{equation}
%	MAP = \frac{1}{n} \sum_{i=1}^{n} AP_i.
%\end{equation}

\begin{table*}[ht!]
\small
\centering
\caption{MAP of Hamming ranking w.r.t. different numbers of bits on VOC2007, Standford Dogs and SUN397 datasets.}
\label{MAP}
\begin{tabular}{|c|cccc|cccc|cccc|}
\hline
\multirow{2}{*}{{Methods}}& \multicolumn{4}{c|}{VOC2007(MAP)}&\multicolumn{4}{c|}{Stanford Dogs(MAP)} & \multicolumn{4}{c|}{SUN397 (MAP)}\\  
& 16 bits & 32 bits & 48 bits &  64 bits& 16 bits& 32 bits & 48 bits&  64 bits & 16 bits & 32 bits & 48 bits &  64 bits \\
\hline
Ours&  \textbf{0.7672}&  \textbf{0.8041}&  \textbf{0.8147}& \textbf{0.8227}&   \textbf{0.6745}&   \textbf{0.7101}& \textbf{0.7252}& \textbf{0.7293} & \textbf{0.7505}& \textbf{0.8068}& \textbf{0.8152}& \textbf{0.8301}\\ \hline

TripletH&0.7434&  0.7745&  0.7858&  0.7897&  0.5889&  0.6478&  0.6744& 0.6904&  0.6893&  0.7660& 0.7918&  0.8027 \\ \hline

DSH&     0.7061&  0.7234& 0.7143&  0.7116&  0.4994&  0.6134&  0.6420&  0.6453&  0.5999&  0.7031& 0.7545&  0.7725 \\ \hline

%CNNH&    0.&  0.&  0.&  0.&   0.&  0.&  0.&  0. \\ \hline 

%KSH&     0.&  0.&  0.&  0.&   0.6337&  0.7208&  0.&  0.&  0.2505&  0.2736&  0.2932& 0.2995&   0.3296&  0.4524&  0.5161&  0.5583 \\ \hline

MLH&     0.4990&  0.5044&  0.4566&  0.4786&   0.4053&  0.5031&  0.6135&  0.6284&  0.3806&  0.4632&  0.4988&  0.5212 \\ \hline

BRE&     0.6111&  0.6323&  0.6453&  0.6517&   0.2429&  0.3158&  0.3683&  0.3936&  0.2546&  0.3185&  0.3574&  0.3675 \\ \hline

ITQ&     0.5793&  0.5808&  0.5796&  0.5723&   0.3275&  0.4256&  0.4682&  0.4977&  0.2789&  0.3684&  0.3873&  0.4013 \\ \hline

SH&      0.4418&  0.4175&  0.4004&  0.3835&   0.2461&  0.3076&  0.3523&  0.3714&  0.2070&  0.2436&  0.2449&  0.2404 \\ \hline

LSH&     0.2720&  0.3098&  0.3273&  0.3551&   0.0629&  0.1293&  0.1690&  0.2245&  0.0794&  0.0981&  0.1211&  0.1486 \\ \hline

\end{tabular}
\end{table*}

\begin{table}[ht!]
\small
\centering
\caption{MAP of Hamming ranking w.r.t. different numbers of bits on CUB-200-2011 dataset.}
\label{MAP2}
\begin{tabular}{|c|cccc|}
\hline
\multirow{2}{*}{{Methods}}& \multicolumn{4}{c|}{CUB-200-2011(MAP)}\\  
& 16 bits & 32 bits & 48 bits &  64 bits \\
\hline
Ours& \textbf{0.5137}&   \textbf{0.6519}& \textbf{0.6807}& \textbf{0.6949}\\ \hline
 
TripletH& 0.4508&  0.5503&  0.5963& 0.6312 \\ \hline
                                                    
DSH&      0.4374&  0.4933&  0.5553&  0.6073 \\ \hline
                                                    
MLH&      0.1069&  0.1510&  0.1768&  0.2091 \\ \hline
                                                    
BRE&      0.0716&  0.0912&  0.1127&  0.1244 \\ \hline
                                                    
ITQ&      0.0899&  0.1266&  0.1427&  0.1599 \\ \hline
                                                    
SH&       0.0703&  0.0875&  0.1017&  0.1111 \\ \hline
                                                    
LSH&      0.0272&  0.0422&  0.0573&  0.0625 \\ \hline                                

\end{tabular}
\end{table}

\subsection{Experimental Setting}
All deep CNN-based methods, including ours and previous baselines, are based on the same CNN architecture, i.e., GoogLeNet~\cite{GoogleLeNet}. We make the following modifications for hashing problem: 1) the last fully-connected layer is removed since it is for 1,000 classifications, and 2) another fully-connected layer with $q$ dimensional output is added to generate the binary codes. The weights are initialized with the pre-trained GoogleNet model~\footnote{http://dl.caffe.berkeleyvision.org/bvlc\_googlenet.caffemodel} that learns from the ImageNet dataset. These experiments are implemented by using the open source \textit{Caffe} framework. All networks are trained by stochastic gradient descent with 0.9 momentum and 0.0005 weight decay. The base learning rate is 0.001 and it is changed to one tenth of the current value after every 50 epochs. The total epoch is 150 and the batch size is 100.  We implement both our methods and comparison ones for varied hash code lengths, e.g., 16 bits, 32 bits, 48 bits and 64 bits. For fair comparison, the hyper-parameters for all deep-network-based methods are the same, including training iterations, batch sizes and etc. For other non-deep-network-based methods, the input features are also extracted by the same pre-trained GoogleNet model, i.e., the last layer's output 1024 dimensional vector. 

The source code of the proposed method will be made publicly available at the first author's homepage. 

\subsection{Experimental Results}

\subsubsection{Comparison with State-of-the-art Methods} In this set of experiments, we evaluate and compare the performance of the proposed method with several state-of-the-art algorithms. 

LSH~\cite{LSH}, SH~\cite{SH}, ITQ~\cite{ITQ}, MLH~\cite{MLH}, BRE~\cite{BRE}, triplet hashing (TripletH)~\cite{onestep} and DSH~\cite{liu2016deep} are selected as the baselines. TripletH is one of the  representative triplet-based methods and DSH is one of the representative pairwise-based methods. The results of these comparison methods are carefully obtained by the implementations provided by their authors, respectively. In TripletH, we replace the original divide-and-encode structure with the fully connected layer to generate hash codes. 

%Note that the most similar work is TriH. TriH and our methods use the same network and the only different is that using or not using the proposed order-aware hard triplet mining, these comparisons can show us whether the proposed order-aware weighting and the squared triplet ranking loss can contribute to the accuracy or not.

Table~\ref{MAP} and Table~\ref{MAP2} show the comparison results of MAP on the four datasets. It can be observed that the proposed method performs significantly better than all previous methods. Specifically, on VOC2017, our method obtains a MAP of 0.8227 on 64 bits, compared with 0.7897 of the existing triplet based method. On Stanford Dogs, our method shows an increase of 2\% in comparison with the TripletH. Figure~\ref{pr-result} and Figure~\ref{p-result} show the precision-recall and precision curves on 16 bits. Again, for most levels, our method yields the better accuracy. The results show that our proposed method can achieve better performance than the existing state-of-the-art methods.

\subsubsection{Effects of the Order-aware Weight and Squared Triplet Loss}
In the second set of our experiments, we do ablation study to clarify the impact of each part of our method on the final performance.

In the first baseline, we only explore the effect of the order-aware weighting factors. The loss function is formulated as
\begin{equation}
\min \sum \lambda_{(i,j,k)} \ell_{(i,j,k)}. 
\end{equation}

In the second baseline, we set the order-aware weighting factors to be one for all triplets, e.g., $\lambda_{i,j,k} = 1$, and only explore the effects of the squared ranking loss. The objective is defined as 
\begin{equation}
\min \sum [\ell_{(i,j,k)}]^2. 
\end{equation}

The last baseline is the existing triplet hashing (TripletH), which the objective is formulated as 
\begin{equation}
\min \sum \ell_{(i,j,k)}. 
\end{equation}

Note that all baselines and our method use the same network and the only difference is the loss function, these comparisons can show us whether the proposed order-aware weights and the squared triplet loss can contribute to the accuracy or not.

\begin{table}[t]
\small
    \centering \caption{Quantitative ablation study on four databases.}
    \begin{tabular}{| c | c c c c |}
        \hline
 Methods & 16 bits & 32 bits & 48 bits & 64 bits \\
        \hline
  \multicolumn{5}{|c|}{VOC2007} \\
       \hline
   Ours &  0.7672&  0.8041&  0.8147 & 0.8227\\ 
   \hline

   Order-aware Weight & 0.7540  &  0.7851 &  0.7975&  0.8052 \\ 
   \hline

   Squared Loss & 0.7552& 0.7997&  0.8049& 0.8160 \\ 
   \hline

   TripletH & 0.7434&  0.7745&  0.7858&  0.7897 \\
   \hline
   \hline
   
  \multicolumn{5}{|c|}{Stanford Dogs} \\
    \hline
   Ours & 0.6745 &  0.7101& 0.7252& 0.7293 \\ 
   \hline

   Order-aware Weight & 0.6548&  0.7058&  0.7291& 0.7368 \\  
   \hline

   Squared Loss & 0.6261 & 0.6891 &  0.6899& 0.7090 \\ 
 \hline

      TripletH & 0.5889&  0.6478&  0.6744&  0.6904 \\
   \hline
   \hline
   
   \multicolumn{5}{|c|}{SUN397} \\
     \hline
    Ours &  0.7505& 0.8068& 0.8152& 0.8301 \\ 
   \hline

   Order-aware Weight &  0.7117& 0.7768& 0.8065& 0.8170 \\  
   \hline

   Squared Loss & 0.7293& 0.7732& 0.8042& 0.8191 \\ 
   \hline

   TripletH &  0.6893&  0.7660& 0.7918&  0.8027 \\
   \hline
   \hline
   
     \multicolumn{5}{|c|}{CUB-200-2011} \\
     \hline
   Ours &  0.5137 & 0.6519 & 0.6807 & 0.6949 \\ 
   \hline

   Order-aware Weight & 0.5169&  0.6158&  0.6507&  0.6779 \\  
   \hline

   Squared Loss & 0.4698 & 0.5723 & 0.6371 & 0.6754 \\ 
   \hline

   TripletH & 0.4508&  0.5503&  0.5963& 0.6312 \\
   \hline
        \end{tabular}
    \label{squared_order}
\end{table} 

Table~\ref{squared_order} shows the comparison results. We can observe that using both the order-aware weights and squared ranking loss performs best. And the proposed order-aware weight and squared loss perform better than the TripletH. It is desirable to reweight the triplets for triplet-based hashing networks.

\subsubsection{Comparison with Hard Triplet Selection Methods} Our method is an order-aware method for reweighting the triplets. To show the advantages of the proposed method, we compare it to the hard triplet selection methods.

The first baseline is hard negative mining (HNM)~\cite{wang2015unsupervised} for triplet sampling. It includes two steps: 1) random selection. They firstly randomly sample the triplets. After 10 epochs of training using data selected randomly, they do 2) hard negative mining, where it selects the top 4 negative triplets with highest losses for each anchor-positive pair. Similar to that, we first use the TripletH to train a hash model (random selection), then we fine-tune the network by using hard negative mining. In each mini-batch, we select the top 4 negative triplets as the suggestion by HNM. The second baseline is semi-hard triplet selection~\cite{schroff2015facenet}. It uses all anchor-positive pairs in a mini-batch and selects the negative examplars that are further away from the anchor than the positive examplar.  

\begin{table}[t]
\small
    \centering \caption{Comparison results of our method against hard triplet selection methods on four datasets.}
    \begin{tabular}{| c | c c c c |}
        \hline
 Methods & 16 bits & 32 bits & 48 bits & 64 bits \\
        \hline
  \multicolumn{5}{|c|}{VOC2007} \\
       \hline
   Ours &  0.7672&  0.8041&  0.8147 & 0.8227\\ 
   \hline

   HNM &  0.7545& 0.7768&  0.8010& 0.8030 \\ 
   \hline
   
   Semi-hard & 0.7347&  0.7442&  0.7548&  0.7655 \\
   \hline

   TripletH & 0.7434&  0.7745&  0.7858&  0.7897 \\
   \hline
   \hline
  \multicolumn{5}{|c|}{Stanford Dogs} \\
    \hline
   Ours & 0.6745 &  0.7101& 0.7252& 0.7293 \\ 
   \hline

   HNM & 0.5905 &  0.6763 &  0.7057 &  0.7187 \\ 
   \hline
   
   Semi-hard & 0.5933&   0.6662&  0.6967&  0.7054 \\
   \hline

      TripletH & 0.5889&  0.6478&  0.6744&  0.6904 \\
   \hline
   \hline
      \multicolumn{5}{|c|}{SUN397} \\
     \hline
    Ours &  0.7505& 0.8068& 0.8152& 0.8301 \\ 
   \hline

   HNM &  0.7124& 0.7801& 0.7992&  0.8142 \\  
   \hline
   
   Semi-hard &  0.7167&  0.7809&  0.8002&  0.8097 \\
   \hline
   
   TripletH &  0.6893&  0.7660& 0.7918&  0.8027 \\
   \hline
   \hline
     \multicolumn{5}{|c|}{CUB-200-2011} \\
     \hline
   Ours &  0.5137 & 0.6519 & 0.6807 & 0.6949 \\ 
   \hline

   HNM & 0.4834 & 0.5793 & 0.6217 & 0.6641 \\ 
   \hline
   
   Semi-hard & 0.4859&   0.6017& 0.6346&  0.6539 \\
   \hline

   TripletH & 0.4508&  0.5503&  0.5963& 0.6312 \\
   \hline
        \end{tabular}
    \label{triplet_selection}
\end{table} 
Table~\ref{triplet_selection} shows the comparison results w.r.t. the MAP on the four datasets. We can see that the proposed method performs better than the loss-based hard triplet mining methods. The results show that order relations can further improve the performance.

\subsubsection{Effects of Different Functions for Triplet Loss}
In this paper, we use the squared triplet loss function to replace the linear form. In this set of experiments, we explore the effects of different functions. In general, the triplet loss can be written into more general form:
 \begin{equation}
 \min \sum [\ell_{(i,j,k}]^\gamma.
 \end{equation}
When $\gamma=1$, it equals to the traditional triplet ranking loss, when $\gamma=2$, it is the squared triplet loss, etc.

\begin{figure}
  \centering
    \includegraphics[width=0.7\hsize \hspace{0.01\hsize}]{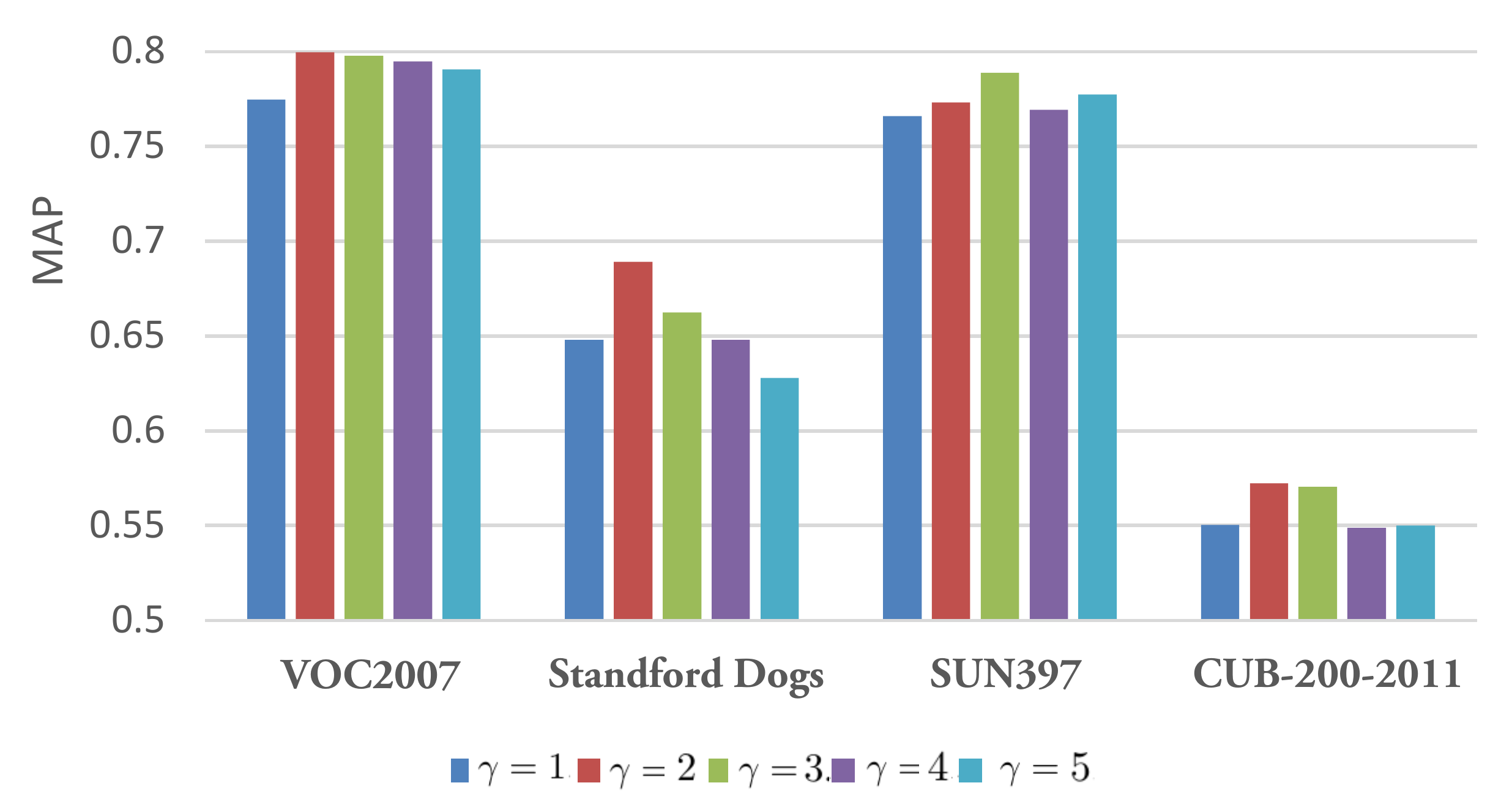}
  \caption{MAP of Hamming ranking w.r.t. different $\gamma$}
  \label{map_variant_gamma}  % be used to cite the whole fig
\end{figure}

Figure~\ref{map_variant_gamma} shows the comparison results, which is implemented on 32 bits, on different functions: $\gamma = 1,\cdots,5$. We can observe that the best results are obtained when $\gamma = 2$ or $\gamma = 3$. Hence, we use the squared triplet ranking loss in the paper.

\section{Conclusion}
In this paper, we proposed an order-aware reweighted method for training the triplet-based deep binary embedding networks. In the proposed deep architecture, images go through the deep network with stacked layers and are encoded into the binary codes. Then, we proposed to up-weight the informative triplets and down-weight the easy triplets by considering the order relations. One is the order-aware weighting factor, which is used to calculate the importance of the triplets. Another is the squared triplet ranking loss, which is used to put more weight on the triplets in which the codes are misranked. Empirical evaluations on four datasets show that the proposed method achieves better performance than the state-of-the-art baselines.

%\section*{Acknowledgments}

\appendix

%\section{Order-aware Weighting factor for MAP}\label{stylefiles}
%
%\textbf{Evaluation Measures}. MAP is a widely used evaluation measure for ranking.  The average precision of image $x_i$ is defined as the average of all
%$P_i@k (k=1,\cdots,c)$ and can be calculate by
%\begin{equation}
%	 AP_i = \sum_{k=1}^c P_i@k \times pos(k)/ |c_{+}^i|,
%\end{equation}
%where $k$ is the position, $c$ is the number of labels, and $pos(k)$ is an indicator function, in which if the label at position $k$ is positive, then $pos(k) = 1$, or else $pos(k) = 0$.
%
%For all images, the MAP is defined as
%\begin{equation}
%	MAP = \frac{1}{n} \sum_{i=1}^{n} AP_i.
%\end{equation}
%
%For MAP, we define $r(j)$ as  the rank of $j$-th label. And we denote $\Delta_b (j, l)$ as the change in MAP that exchange the ranks of the $j$-th and $l$-th labels. For ease representation, we assume that $r(j) < r(l)$ and $b$ is the number of positive labels ranked topper than $j$-th label.
% 
%For $r(l) = r(j) + 1 $, we have
%\begin{equation}
%  \Delta_b (j, l) = \frac{1}{|c_{+}|} ( \frac{b}{b+r(j)} - \frac{b-1}{b+r(j)-1} )
%\end{equation}
%and hence $\Delta_b (j, k)$ can be calculated by sum of the pairs $(r(j) ,r(j)+1),\cdots,(r(l)-1,r(l))$.

\bibliographystyle{named}
\bibliography{paper}

\end{document}